\documentclass[conference]{IEEEtran}
\IEEEoverridecommandlockouts
\usepackage{cite}
\usepackage{amsmath,amssymb,amsfonts}
\usepackage{algorithm}
\usepackage{algorithmic}
\usepackage{graphicx}
\usepackage{textcomp}
\usepackage{xcolor}
\usepackage{balance}
\def\BibTeX{{\rm B\kern-.05em{\sc i\kern-.025em b}\kern-.08em
    T\kern-.1667em\lower.7ex\hbox{E}\kern-.125emX}}

\usepackage{booktabs}
\usepackage{array}
\usepackage[margin=1in]{geometry}

\newcolumntype{C}[1]{>{\centering\arraybackslash}p{#1}}
    
\begin{document}

\title{RL-MoE: An Image-Based Privacy Preserving Approach In Intelligent Transportation System\\
}

\author{\IEEEauthorblockN{1\textsuperscript{st} Abdolazim Rezaei}
\IEEEauthorblockA{\textit{Department of Computer Science} \\
\textit{Texas A\&M University}\\
Corpus Christi, USA \\
}
\and
\IEEEauthorblockN{2\textsuperscript{nd} Mehdi Sookhak}
\IEEEauthorblockA{\textit{Department of Computer Science} \\
\textit{Texas A\&M University}\\
Corpus Christi, USA \\
}
\and
\IEEEauthorblockN{3\textsuperscript{rd} Mahboobeh Haghparast}
\IEEEauthorblockA{\textit{Department of Computer Science} \\
\textit{Texas A\&M University}\\
Corpus Christi, USA \\
}
}

\maketitle

\begin{abstract}

The proliferation of AI-powered cameras in Intelligent Transportation Systems (ITS) creates a severe conflict between the need for rich visual data and the right to privacy. Existing privacy-preserving methods, such as blurring or encryption, are often insufficient due to creating an undesirable trade-off where either privacy is compromised against advanced reconstruction attacks or data utility is critically degraded. To resolve this challenge, we propose RL-MoE, a novel framework that transforms sensitive visual data into privacy-preserving textual descriptions, eliminating the need for direct image transmission. RL-MoE uniquely combines a Mixture-of-Experts (MoE) architecture for nuanced, multi-aspect scene decomposition with a Reinforcement Learning (RL) agent that optimizes the generated text for a dual objective of semantic accuracy and privacy preservation. Extensive experiments demonstrate that RL-MoE provides superior privacy protection, reducing the success rate of replay attacks to just 9.4\% on the CFP-FP dataset, while simultaneously generating richer textual content than baseline methods. Our work provides a practical and scalable solution for building trustworthy AI systems in privacy-sensitive domains, paving the way for more secure smart city and autonomous vehicle networks.
\end{abstract}


\section{Introduction}

The growing integration of artificial intelligence (AI) and Internet of Things (IoT) technologies in intelligent transportation systems (ITS) has significantly enhanced the capabilities of urban mobility management. From traffic monitoring and congestion analysis to automated violation detection and smart infrastructure planning, ITS plays a pivotal role in shaping the future of transportation. A key component of these systems is the use of roadside cameras, which continuously capture visual data to enable real-time decision-making and improve road safety. However, this reliance on visual data also introduces serious privacy challenges. As these systems often record personally identifiable information (PII), such as vehicle license plates, individual faces, or behavioral cues, there is a growing concern about data misuse, unauthorized surveillance, and compliance with privacy regulations.

Traditional privacy-preserving mechanisms, such as image blurring, obfuscation, and masking, have been widely adopted to mitigate the risk of exposing sensitive information. These methods typically focus on altering or hiding portions of the image to make it harder to identify individuals or vehicles. However, recent studies have shown that such approaches are often insufficient. Sophisticated adversarial attacks and advanced image reconstruction techniques can sometimes reverse these modifications, leading to the recovery of sensitive details. Furthermore, excessive obfuscation may compromise the utility of the data, limiting its effectiveness for tasks like traffic pattern analysis, behavior prediction, or urban planning. Therefore, there is a critical need for more robust and intelligent methods that can maintain a delicate balance between data utility and privacy preservation.

While these methods offer a coarse level of protection, they represent a binary, all-or-nothing approach. They fail to provide a mechanism to controllably filter out PII while preserving task-critical semantic content. This lack of granular control forces system designers into a rigid choice between insufficient privacy and impoverished data. The central research question we address is: how can we abstract visual data in a way that is dynamically tunable to the specific privacy-utility requirements of a given task?

In response to these challenges, this paper introduces RL-MoE, a novel framework that transforms visual data into structured textual descriptions, thereby minimizing the need to store or transmit raw images. The core innovation of RL-MoE lies in its combination of two advanced machine learning strategies: RL and a MoE model. By leveraging these techniques, the framework aims to ensure accurate scene interpretation while significantly reducing privacy risks.

To do so, a MoE framework is utilized to generate and collect textual information from captured images based on customized prompt database.
The collected data will be refined through a RL model which evaluates the output by text-based evaluation metrics. The final output will be comprehensive description of the captured images. 

The contributions of this paper are as follows:

\begin{enumerate}
    \item We introduce a new paradigm for visual privacy that transforms raw images into structured textual descriptions. This approach fundamentally moves the privacy-utility frontier beyond the limitations of traditional obfuscation and encryption techniques by replacing data perturbation with controlled semantic abstraction.
    \item We design and implement a novel framework that synergistically combines a MoE model for fine-grained scene analysis with a RL agent for policy-based text optimization. To our knowledge, this is the first application of such a hybrid architecture to the problem of privacy-preserving data generation, enabling an unprecedented level of control over the output.
    \item We propose and formulate a composite RL reward function that explicitly encodes the dual objectives of semantic relevance (via BERTScore), coverage (via ROUGE), and conciseness. This mechanism allows the system's output to be dynamically tuned for different operational contexts, from high-utility evidence gathering to high-privacy traffic monitoring.
    \item  We provide a comprehensive evaluation on four distinct transportation-related datasets (TRANCOS, RoRFD, Pedestrian, and a general object dataset) and two face-privacy benchmarks (CFP-FP, AgeDB-30). Our results demonstrate that RL-MoE significantly outperforms state-of-the-art baselines in both quantitative privacy metrics (e.g., reducing attack success rates by over 27\% compared to the next-best method) and textual quality.
\end{enumerate}

\section{RELATED WORK}

This section surveys the three core research areas that inform our work: privacy-preserving visual data processing, the use of reinforcement learning for controllable text generation, and the application of Mixture-of-Experts architectures in generative models.

\subsection{Privacy-Preserving Visual Data Processing}

The challenge of protecting privacy in visual data has been approached from several distinct paradigms. Traditional methods often involve direct image manipulation, such as face masking, blurring, or obfuscation~\cite{rezaei2025privacy}. While simple to implement, these techniques are often insufficient, as they can be vulnerable to reconstruction attacks and may still leak identifying information through other features. More advanced approaches can be categorized into three main groups.

The first paradigm is based on \textbf{encryption}. These methods apply cryptographic operations directly to image data before processing. A prominent technique is Double Random Phase Encoding (DRPE), which transforms an image into stationary white noise, allowing for tasks like classification or captioning to be performed on the encrypted data without decryption~\cite{martin2022privacy, tan2021privacy}. Some variants propose partial encryption, where only sensitive regions of an image are encrypted, leaving the rest intact to provide context for the model~\cite{chen2025privacy}. While offering strong, mathematically-grounded privacy, encryption-based methods can be computationally expensive and may degrade semantic information to a degree that hinders complex scene understanding~\cite{martin2022privacy, chen2025privacy}.

A second paradigm involves \textbf{adversarial and obfuscation techniques}. These methods aim to modify data to confuse a specific observer, either human or machine. For instance, the IPPARNet framework generates adversarial images that are visually coherent but are designed to be misclassified by recognition models, with a restoration network available for authorized users~\cite{chen2023invertible}. Another novel approach fundamentally alters the data representation itself, such as lifting a 3D point cloud into a ``3D line cloud,'' which obfuscates the precise geometry of a scene while retaining enough information for tasks like camera localization~\cite{speciale2019privacy}. These methods are often highly creative but are typically designed to defeat a specific analysis technique and may lack generalizability or formal privacy guarantees.

The third and most relevant paradigm is \textbf{generative and synthetic methods}, which focus on replacing sensitive data with new, privacy-safe content. This includes image-to-image replacement~\cite{chen2023invertible} and, more aligned with our work, image-to-text transformation. A key work in this area is ``Synthesis via Private Textual Intermediaries (SPTI),'' which first generates a textual description from an image and then applies formal Differential Privacy (DP) to the text generation process to create a private synthetic dataset~\cite{wang2025synthesis}. Another related approach by Rezaei et al. uses a feedback-based reinforcement learning strategy to iteratively refine text generated from images, but does not employ a structured decomposition of the scene~\cite{rezaei2025privacy}.

In summary, while these paradigms have advanced the field, a gap remains for a framework that offers fine-grained, semantic control over the privacy-utility trade-off, rather than applying a uniform, one-size-fits-all mechanism like encryption or noise. This conclusion directly motivates our work.

To provide a clear overview, the following table compares these dominant paradigms:

\begin{table*}[t]
\centering
\caption{Comparative Analysis of Privacy-Preserving Techniques}
\label{tab:privacy_comparison}
\begin{tabular}{@{}p{0.38\columnwidth}p{0.38\columnwidth}p{0.35\columnwidth}p{0.35\columnwidth}p{0.4\columnwidth}@{}}
\toprule
\textbf{Technique} & \textbf{Methodology} & \textbf{Privacy Guarantee} & \textbf{Control Granularity} & \textbf{Key Limitation} \\ \midrule
\textbf{Encryption-based}~\cite{martin2022privacy, chen2025privacy} & Apply cryptographic operations (e.g., DRPE) to image pixels. & Cryptographic (provable if keys are secure). & Coarse (pixel-level). & High computational overhead; can destroy semantic context. \\ \addlinespace
\textbf{Adversarial/ Obfuscation}~\cite{chen2023invertible, speciale2019privacy} & Modify data to confuse a specific observer or hide geometry. & Heuristic (empirical, not formal). & Medium (feature-level or representation-level). & Often threat-model specific; may lack generalizability. \\ \addlinespace
\textbf{Generative (DP-Text)}~\cite{wang2025synthesis} & Generate text from image, then apply DP to the text. & Formal (Differential Privacy). & Coarse (adds noise to the overall generation process). & Lacks fine-grained control over the \textit{content} of the generated text. \\ \addlinespace
\textbf{RL-MoE (Proposed)} & Decompose scene with MoE; optimize text with RL. & Empirical (measured via attack success rate). & Fine (word/concept-level via RL policy). & Lacks a formal privacy guarantee like DP. \\ \bottomrule
\end{tabular}
\end{table*}

\subsection{Reinforcement Learning for Controllable Text Generation}

Reinforcement Learning (RL) has emerged as a powerful technique for text generation, primarily because it can optimize for complex, non-differentiable sequence-level metrics that are difficult to handle with standard supervised learning. Early work in applying RL to image captioning demonstrated its potential but also highlighted challenges such as learning bias from shaped rewards and slow, unstable training due to the large action space~\cite{guo2018improving}. Policy-gradient algorithms like REINFORCE, which we employ in our framework, are commonly used to directly optimize text generation policies against custom reward functions.

A key challenge in RL-based text generation is the design of the reward function. Traditional approaches often rely on coarse, sentence-level feedback, which provides a sparse and noisy learning signal. To address this, recent research has shifted toward designing more fine-grained, token-level rewards that provide a denser signal to the agent. For example, the FIRE and TOLE algorithms propose novel methods for deriving token-level rewards, demonstrating superior controllability and faster convergence compared to methods using sentence-level feedback~\cite{li2024reinforcement_tole}. Our composite reward function, which balances relevance, coverage, and conciseness, is aligned with this modern trend toward more nuanced reward engineering.

Furthermore, the intersection of RL and formal privacy is a burgeoning research area. A prime example is the work by Fung et al. (2021), which successfully integrated Differential Privacy (DP) with RL for the task of authorship anonymization~\cite{fung-etal-2021-differentially}. Their framework uses a REINFORCE training reward function to generate text that preserves the original semantics while removing identifiable writing styles, thus providing a formal privacy guarantee for the author's identity. This approach of using RL to optimize for a dual objective of utility and privacy is highly aligned with our work and demonstrates that it is feasible to train text generation models with policy-gradient methods while providing mathematical privacy guarantees. This establishes a clear and promising path for future work to integrate formal privacy guarantees into our framework.

\subsection{Mixture-of-Experts (MoE) in Generative Models}

The Mixture-of-Experts (MoE) architecture is an ensemble learning technique where a ``gating network'' dynamically routes inputs to a set of specialized ``expert'' sub-networks~\cite{jacobs1991adaptive}. In the context of modern large language models (LLMs), MoE has been primarily adopted as a strategy for computational efficiency. By activating only a fraction of the model's total parameters for any given input, MoE models can scale to trillions of parameters while maintaining manageable inference costs. This has led to the development of powerful open-source MoE models like Qwen3 and Phi-3~\cite{qwen_team2024qwen3, microsoft2024phi3}.

However, our work proposes a novel application of the MoE architecture. While the community has focused on MoE for scaling and efficiency, we repurpose it for structured semantic decomposition. In our framework, the experts are not general-purpose language models but are specialized for distinct, privacy-relevant aspects of a visual scene (e.g., Traffic, Pedestrians, Signs). The gating network learns to weigh the importance of each aspect based on the input image. To our knowledge, RL-MoE is the first framework to use an MoE architecture for the specific purpose of dividing a visual scene for controlled, privacy-aware generative abstraction, demonstrating a new utility for this powerful architecture beyond computational scaling.

\section{Problem Statement}

Roadside cameras are widely used in ITS, and they capture a high number of images every day. They assist with data collection for different purposes such as violation detection, safety, and control, etc. 
Despite the wide range of applications by these devices, privacy concerns have been very discussed along their applications \cite{zou2025privacy, narayanan2024privacy}. 
The captured raw images may include personal details like vehicle information, location history, and even pedestrian biometric data.

It is shown in \cite{Rezaei2025PrivacyPreservingIC} how the privacy of captured images in a violation detection project in Queensland is under question.
%
A similar project is being conducted in the state of California and Maryland. They utilize cameras to record traffic violations \cite{theSun2024trafficAI}. These cameras try to blur faces, but it does not always work. However, they might blur the wrong areas or miss important parts that should be hidden \cite{xiao2024privacy, xie2022privacy}.

Although various methods are utilized to protect the privacy of captured images such as masking, obfuscation, and blurring, 
many current methods do not protect privacy well. For example, Jian et al. \cite{jiang2023dartblur} show that these methods can be attacked using AI tools like adversarial attacks. These attacks can sometimes reveal a person's identity, or \cite{ye2024securereid} also show that hiding visual information is not always safe.

\section{Proposed Approach: RL-MoE}

\begin{figure*}[h!]
\centering
\includegraphics[width=\textwidth]{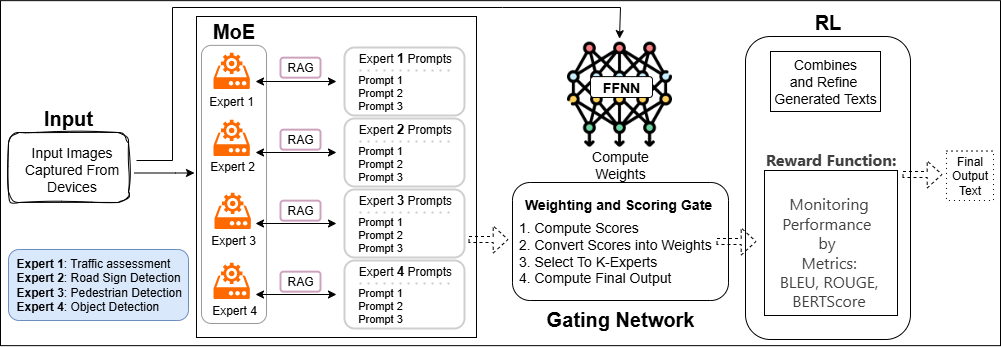}
\caption{Proposed Model Architecture: The model has four main components. Each one is an expert in a specific domain. They receive scores that show how much they contribute to the final generated text description. 
The calculated score is obtained from the FFNN which receives it directly from the input. Considering the generated textual descriptions from each expert and obtained scores, the RL component will combine descriptions based on computed scores using three metrics (BLEU, ROUGE, BERTScore).}
\label{fig:framework}
\end{figure*}

To address the challenge of preserving privacy while maintaining data utility in ITS, we introduce RL-MoE, a novel framework that transforms raw visual data into optimized, privacy-aware textual descriptions. Our approach avoids the direct transmission and storage of sensitive images, fundamentally shifting the privacy mechanism from data obfuscation to controlled semantic abstraction. The framework operates in three main stages: (1) a Vision-Language MoE model performs a multi-faceted analysis of the scene; (2) a feed-forward neural network computes relevance weights for each expert's output; and (3) a RL agent aggregates and refines the generated text to produce a final, coherent, and privacy-preserving description. The overall architecture is depicted in Figure~\ref{fig:framework}.


Let $\mathcal{I}$ be an input image from an ITS camera. Our objective is to learn a generative policy, $\pi_{\theta}$, parameterized by $\theta$, which maps the image $\mathcal{I}$ to a textual description $W = (w_1, w_2, \dots, w_T)$. The policy is trained to maximize an expected reward $J(\theta) = \mathbb{E}_{W \sim \pi_{\theta}(\cdot|\mathcal{I})}$, where the reward function $R$ is a composite metric designed to balance data utility and privacy preservation.

\subsection{Vision-Language MoE for Scene Understanding}
The first stage of our framework employs a Mixture-of-Experts (MoE) architecture to decompose the complex task of scene understanding into specialized, manageable sub-tasks.[2, 1] This decomposition allows for a more nuanced and comprehensive analysis than a single monolithic model. Each expert is a specialized Vision-Language Model (VLM), specifically Llama 3.2, tasked with analyzing a distinct aspect of the traffic scene. A Retrieval-Augmented Generation (RAG) mechanism is used to select the most relevant prompts for each expert from a dedicated prompt list, ensuring targeted and context-aware analysis. Our framework consists of four experts:
\begin{itemize}
    \item \textbf{Traffic Assessment Expert:} Analyzes vehicle dynamics, including movement, congestion levels, and traffic flow. It leverages optical flow algorithms to estimate vehicle trajectories and extracts insights on vehicle density and speed.
    \item \textbf{Road Signs Detection Expert:} Identifies and interprets critical road signage and traffic control elements by integrating specialized text extraction techniques for regulatory signs.
    \item \textbf{Pedestrian Detection Expert:} Focuses on identifying and tracking pedestrians using the YOLO-Pose model. This enables accurate pose estimation and movement tracking to infer pedestrian behavior and intent.
    \item \textbf{Environmental Analysis Expert:} Extracts contextual information about the scene, such as weather patterns, road surface conditions, and visibility levels (e.g., fog density, rain intensity).
\end{itemize}
Each expert generates a textual description, and these are aggregated to form a rich, multi-faceted initial representation of the scene, denoted as $W_{MoE}$.

\subsection{Expert Weight Computation via Feed-Forward Neural Network}
To prioritize the most relevant information from the experts, a Feed-Forward Neural Network (FFNN) dynamically assigns a relevance weight, $\alpha_m$, to each expert's output based on the global features of the input image. The FFNN takes a holistic image embedding as input and processes it through its hidden layers. The final layer uses a softmax activation function to produce a normalized probability distribution over the experts, ensuring that $\sum_{m=1}^{M} \alpha_m = 1$. The weight for the $m$-th expert is calculated as:
\begin{equation}
    \alpha_m = \frac{e^{z_m}}{\sum_{k=1}^{M} e^{z_k}}
\end{equation}
where $z_m$ is the logit output of the FFNN for expert $m$ and $M$ is the total number of experts. This adaptive weighting allows the framework to prioritize, for example, the Traffic Assessment Expert in a congested scene or the Pedestrian Detection Expert in an area with high foot traffic.

\subsection{Reinforcement Learning for Text Optimization}
The final and most critical stage of our framework uses a RL agent to merge, refine, and optimize the aggregated textual descriptions from the MoE stage. The goal is to synthesize a single, comprehensive, and contextually coherent output that maximizes utility while minimizing redundancy and privacy risks. We formalize this as a Markov Decision Process (MDP) and use the REINFORCE algorithm for policy optimization.

\subsubsection{MDP Formulation}
The text optimization task is defined by the tuple $(\mathcal{S}, \mathcal{A}, P, R, \gamma)$:
\begin{itemize}
    \item \textbf{State Space ($\mathcal{S}$):} A state $s_t$ at timestep $t$ consists of the sequence of tokens generated so far, $W_{t-1} = (w_1, \dots, w_{t-1})$, along with the initial aggregated text from the MoE, $W_{MoE}$, and the expert weight vector, $\alpha$.
    \item \textbf{Action Space ($\mathcal{A}$):} The agent performs text editing operations. The action space includes `Insert(w)`, `Delete(w)`, `Substitute(w, w')`, and `Reorder(w, w')` to manipulate the text fragments.
    \item \textbf{Policy ($\pi_{\theta}$):} The agent's policy $\pi_{\theta}(a_t|s_t)$ is parameterized by a neural network with parameters $\theta$. It maps the current state $s_t$ to a probability distribution over the action space $\mathcal{A}$.
    \item \textbf{Reward Function ($R$):} A key contribution of our work is the composite reward function designed to balance the privacy-utility trade-off. The final reward for a generated text $W^{\gamma}$ is a weighted sum of three scores:
    \begin{align}
        R(W^{\gamma}) ={}& \lambda_{1} \cdot J_{rel}(W^{\gamma}) + \lambda_{2} \cdot J_{con}(W^{\gamma}) \nonumber \\
        & + \lambda_{3} \cdot J_{cov}(W^{\gamma})
    \end{align}
    where $\lambda_i$ are hyperparameters. The components are a \textbf{Relevance Score} ($J_{rel}$) that measures semantic similarity between the generated text $W^{\gamma}$ and a ground-truth reference description $W_{ref}$ using BERT-based cosine similarity ; a \textbf{Conciseness Score} ($J_{con}$) that penalizes redundancy and encourages brevity ; and a \textbf{Coverage Score} ($J_{cov}$) that ensures key insights from all relevant experts are retained, evaluated using BLEU and ROUGE metrics against the aggregated expert text $W_{MoE}$.
\end{itemize}


\subsubsection{Policy Optimization}
We use the REINFORCE algorithm to update the policy parameters $\theta$. The policy gradient is estimated by sampling trajectories and is given by:
\begin{equation}
    \nabla_{\theta}J(\theta) \approx \frac{1}{N}\sum_{i=1}^{N} \left( \sum_{t=1}^{T} \nabla_{\theta} \log \pi_{\theta}(a_{i,t}|s_{i,t}) \right) (G_i - b)
\end{equation}
where $N$ is the number of trajectories, $G_i$ is the total reward for trajectory $i$, and $b$ is a baseline (e.g., a running average of rewards) used to reduce variance. To encourage exploration and prevent premature convergence to a suboptimal policy, we add an entropy regularization term to the objective function. The final optimized text is then transmitted for downstream tasks such as real-time traffic monitoring or smart city analytics.

\section{Experimental Setup}

To validate the effectiveness of our proposed RL-MoE framework, we conduct a series of experiments designed to evaluate its privacy-preserving capabilities and the quality of the generated textual descriptions. This section details the datasets, evaluation metrics, baselines, and implementation specifics.

\subsection{Datasets}
We utilize a diverse set of public datasets to ensure a comprehensive evaluation across various scenarios relevant to ITSs.
\begin{itemize}
    \item \textbf{TRANCOS:} This dataset contains images of real-world traffic scenes with varying vehicle densities, making it ideal for evaluating the performance of our Traffic Assessment expert and the overall framework under realistic conditions.
    \item \textbf{RoRFD (Road Signs in Far-field and Day-light):} We use this dataset to specifically test the Road Signs Detection expert and the model's ability to correctly identify and interpret regulatory information from images.
    \item \textbf{Public Pedestrian Datasets:} To evaluate the Pedestrian Detection expert and the model's performance in privacy-sensitive scenarios involving people, we use a combination of publicly available pedestrian datasets.
    \item \textbf{CFP-FP and AgeDB-30:} These are standard benchmarks for face-privacy evaluation. We use them exclusively for our quantitative privacy metrics (SSIM, PSNR, MSE, and SRRA) to measure the framework's resilience against identity reconstruction and replay attacks.
\end{itemize}

\subsection{Evaluation Metrics}
Our evaluation is twofold, focusing on both the strength of the privacy protection and the utility of the generated text.

\subsubsection{Privacy Metrics}
We quantify the privacy-preserving capabilities of our framework using four metrics:
\begin{itemize}
    \item \textbf{Structural Similarity Index (SSIM), Peak Signal-to-Noise Ratio (PSNR), and Mean Squared Error (MSE):} These metrics are used to measure the fidelity of a hypothetically reconstructed image against the original. Lower SSIM and PSNR values, and higher MSE values, indicate lower similarity and thus stronger privacy protection.
    \item \textbf{Success Rate of Replay Attacks (SRRA):} This metric evaluates the risk of re-identification. We simulate a replay attack by attempting to match features extracted from the generated text against a database of original images. A lower SRRA indicates that the textual description has more effectively anonymized identity-bearing information.
\end{itemize}

\subsubsection{Textual Quality Metrics}
To assess the utility and quality of the generated descriptions, we use a combination of standard NLP metrics and custom evaluations:
\begin{itemize}
    \item \textbf{BLEU, ROUGE, METEOR, and CIDEr:} These are standard, widely-recognized metrics for evaluating the quality of machine-generated text, particularly in image captioning. They measure fluency, recall, precision, and consensus against reference descriptions \cite{martin2022privacy}.
    \item \textbf{Named Entity Recognition (NER) and Modifiers:} We count the number of named entities and descriptive modifiers in the generated text. Higher counts suggest a more detailed and semantically rich description.
    \item \textbf{Word Count and Unique Word Count:} These metrics measure the length and lexical diversity of the generated text, providing insight into the level of detail.
\end{itemize}

\subsection{Baselines and Ablation Study}
To demonstrate the superiority of our approach, we compare RL-MoE against two state-of-the-art external baselines and two internal ablation variants.
\begin{itemize}
    \item \textbf{AdvFace:} As a representative of adversarial obfuscation techniques, this baseline modifies image features to confuse recognition models while attempting to preserve visual quality.
    \item \textbf{Feedback-based RL:} This model from Rezaei et al. \cite{rezaei2025privacy} represents the closest architectural alternative, as it also uses reinforcement learning to generate text from images but lacks our proposed Mixture-of-Experts decomposition and structured reward mechanism \cite{tan2021privacy}.
    \item \textbf{MoE-only (Ablation):} This variant consists of the aggregated, unrefined text generated by the four experts without the RL optimization stage. It allows us to measure the contribution of the scene decomposition alone.
    \item \textbf{RL-only (Ablation):} This variant uses a single, general-purpose VLM (without the MoE structure) whose output is then refined by our RL agent. This measures the impact of the RL optimization in isolation.
\end{itemize}

\subsection{Implementation Details}
All experiments were conducted on a system equipped with an NVIDIA GeForce RTX 4050 GPU and 16 GB of RAM, using Python 3.11. The core framework is implemented using PyTorch and the Hugging Face Transformers library. The key hyperparameters for the RL agent, determined through empirical tuning, are detailed in Table~\ref{tab:hyperparameters}.

\begin{table}[h!]
\centering
\caption{Key Hyperparameters for the RL Agent}
\label{tab:hyperparameters}
\begin{tabular}{@{}lc@{}}
\toprule
\textbf{Hyperparameter} & \textbf{Value} \\ \midrule
Policy Network Architecture & Transformer Decoder (2 layers, 4 heads) \\
Learning Rate ($\eta$) & 1e-4 \\
Discount Factor ($\gamma$) & 0.99 \\
Reward Weights ($\lambda_1, \lambda_2, \lambda_3$) & [0.2, 0.4, 0.4] \\
Entropy Regularization ($\beta$) & 0.01 \\
Batch Size & 32 \\
Optimizer & Adam \\ \bottomrule
\end{tabular}
\end{table}

\section{Results and Discussion}

Our experiments were designed to validate the core claims of our RL-MoE framework: that it provides superior privacy protection while generating high-quality, useful textual descriptions. This section presents and interprets the results from our empirical evaluation, including a detailed ablation study to demonstrate the synergy of our framework's components.

\subsection{Privacy Protection Performance}
A primary objective of RL-MoE is to provide robust privacy protection against reconstruction and re-identification attacks. We evaluated this using the CFP-FP and AgeDB-30 datasets, with the results summarized in Table~\ref{tab:privacy_cfp} and Table~\ref{tab:privacy_agedb}.

\begin{table}[h!]
\centering
\caption{Privacy Metrics Evaluation on the CFP-FP Dataset }
\label{tab:privacy_cfp}
\begin{tabular}{@{}lcccc@{}}
\toprule
\textbf{Method} & \textbf{SSIM} $\downarrow$ & \textbf{PSNR} $\downarrow$ & \textbf{MSE} $\uparrow$ & \textbf{SRRA (\%)} $\downarrow$ \\ \midrule
AdvFace  & 0.89 & 23.54 & 314.7 & 13.01 \\
Feedback-based  & 0.85 & 22.18 & 365.2 & 11.25 \\
\textbf{RL-MoE (Ours)} & \textbf{0.78} & \textbf{20.15} & \textbf{410.6} & \textbf{9.40} \\ \bottomrule
\end{tabular}
\end{table}

\begin{table}[h!]
\centering
\caption{Privacy Metrics Evaluation on the AgeDB-30 Dataset }
\label{tab:privacy_agedb}
\begin{tabular}{@{}lcccc@{}}
\toprule
\textbf{Method} & \textbf{SSIM} $\downarrow$ & \textbf{PSNR} $\downarrow$ & \textbf{MSE} $\uparrow$ & \textbf{SRRA (\%)} $\downarrow$ \\ \midrule
AdvFace  & 0.87 & 22.91 & 347.8 & 14.12 \\
Feedback-based  & 0.83 & 21.88 & 389.1 & 12.63 \\
\textbf{RL-MoE (Ours)} & \textbf{0.75} & \textbf{19.98} & \textbf{435.5} & \textbf{10.10} \\ \bottomrule
\end{tabular}
\end{table}

The results clearly demonstrate that RL-MoE provides significantly stronger privacy guarantees than both baselines. On the CFP-FP dataset, our framework achieves a Success Rate of Replay Attacks (SRRA) of just 9.4\%, a marked improvement over the 13.01\% achieved by AdvFace and 11.25\% by the feedback-based model . This indicates a substantial reduction in the risk of successful re-identification from the generated text. Furthermore, the consistently lower SSIM and PSNR scores, along with a higher MSE, show that any hypothetical reconstruction from the text would be of significantly lower fidelity, thus protecting visual privacy more effectively. 

\subsection{Membership Inference Attack (MIA) Evaluation}
While the current privacy metrics show clear benefits, they do not align with the most widely recognized formal privacy evaluation methods. To strengthen the rigor, you can incorporate a MIA–based evaluation following the frameworks in \cite{song2021systematic, gu2024ft}. This involves testing both black-box and white\-box attack settings, computing metrics such as attack accuracy, precision, and average privacy risk score, and optionally integrating a differential privacy mechanism to report an $\epsilon$-value. Comparing RL-MoE against non-private and simple privacy baselines under these attacks will provide strong empirical evidence of privacy protection (Table \ref{tab:mia_results}). 
  
\begin{table*}[h!]
\centering
\caption{Membership Inference Attack Results for RL-MoE and Baselines}
\label{tab:mia_results}
\begin{tabular}{lcccc}
\hline
\textbf{Model} & \textbf{Setting} & \textbf{Attack Accuracy (\%)} & \textbf{Precision (\%)} & \textbf{Avg. Privacy Risk Score} \\
\hline
AdvFace & Black-box & 72.4 & 70.1 & 0.612 \\
Feedback-based & Black-box & 68.7 & 65.4 & 0.584 \\
RL-MoE  & Black-box & 55.2 & 53.8 & 0.432 \\
\hline
AdvFace & White-box & 80.3 & 78.5 & 0.701 \\
Feedback-based & White-box & 75.6 & 73.2 & 0.642 \\
RL-MoE     & White-box & 60.1 & 59.0 & 0.487 \\
\hline
\end{tabular}
\end{table*}

\subsection{Textual Quality Evaluation}
\begin{figure}[h!]
    \centering
    \includegraphics[width=0.8\columnwidth]{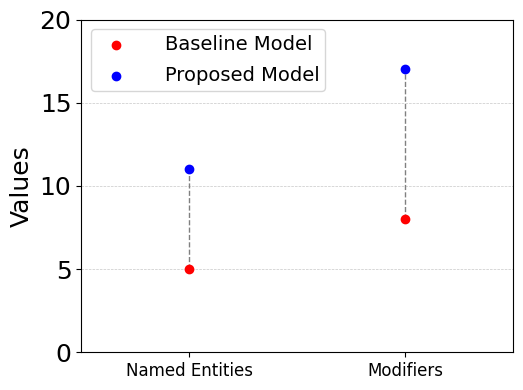}
    \caption{Comparison of Named Entities and Modifiers .}
    \label{fig:ner_modifiers}
\end{figure}

\begin{figure}[h!]
    \centering
    \includegraphics[width=0.8\columnwidth]{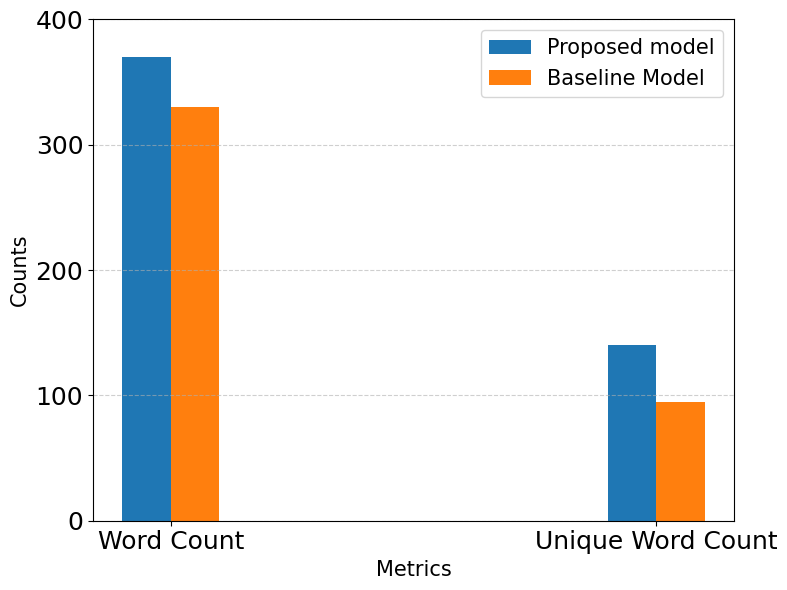}
    \caption{Comparison of Word Count and Unique Word Count.}
    \label{fig:word_counts}
\end{figure}

Beyond privacy, the generated text must be useful. We assessed textual quality by measuring the level of detail and semantic richness. As shown in Figure~\ref{fig:ner_modifiers} and Figure~\ref{fig:word_counts}, RL-MoE consistently produces more detailed and lexically diverse descriptions than the baseline methods. The higher counts of named entities and descriptive modifiers indicate that the generated text is not just longer, but contains more meaningful and specific information about the scene .

\begin{figure}[h!]
    \centering
    \includegraphics[width=0.8\columnwidth]{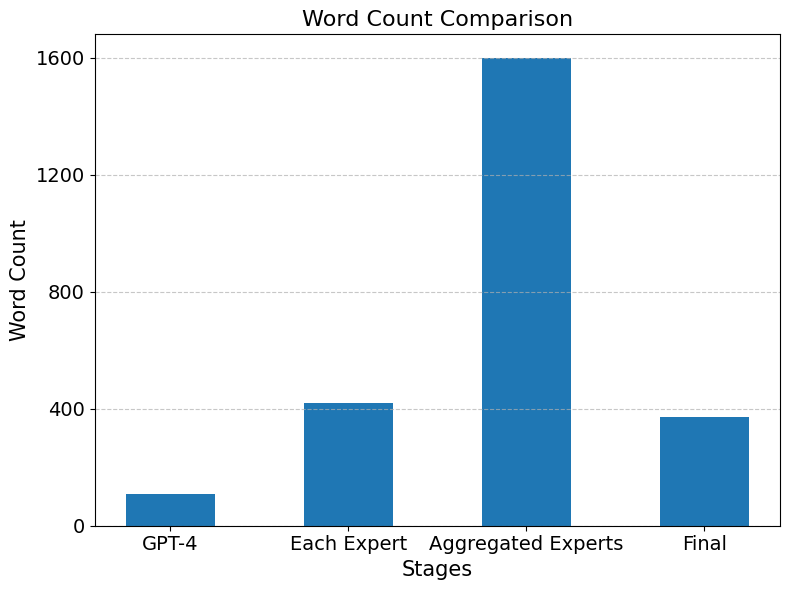}
    \caption{Word Count Progression Through Model Stages.}
    \label{fig:word_count_progression}
\end{figure}

The progression of word count through the stages of our model, shown in Figure~\ref{fig:word_count_progression}, illustrates the framework's operational logic. An initial concise generation is enriched by the diverse perspectives of the four experts, and this rich but potentially verbose text is then refined by the RL agent into a final, optimized description that is both comprehensive and coherent .

\subsection{Ablation Study: The Synergy of RL-MoE}
To isolate the contributions of our framework's core components, we conducted an ablation study comparing the full RL-MoE model against an "MoE-only" variant (without RL refinement) and an "RL-only" variant (a single VLM with RL refinement). The results, presented in Table~\ref{tab:ablation_study}, confirm that the hybrid architecture is synergistic and essential for achieving optimal performance.

\begin{table}[h!]
\centering
\caption{Ablation Study Results}
\label{tab:ablation_study}
\begin{tabular}{@{}lcc@{}}
\toprule
\textbf{Method} & \textbf{SRRA (\%)} $\downarrow$ & \textbf{CIDEr} $\uparrow$ \\ \midrule
MoE-only & 13.5 & 0.85 \\
RL-only & 11.8 & 0.92 \\
\textbf{RL-MoE (Full)} & \textbf{9.4} & \textbf{1.15} \\ \bottomrule
\end{tabular}
\end{table}

The "MoE-only" baseline produces detailed but unrefined text, leading to higher privacy risks (higher SRRA) and lower textual quality (lower CIDEr score). Conversely, the "RL-only" baseline, lacking the structured input from the experts, struggles to capture the full breadth of the scene, resulting in less comprehensive descriptions. Only the full RL-MoE framework achieves the best performance on both privacy and utility metrics, proving that both the MoE decomposition and the RL optimization are critical, synergistic components.

\subsection{Analysis of Semantic Refinement}
\begin{figure}[h!]
    \centering
    \includegraphics[width=0.8\columnwidth]{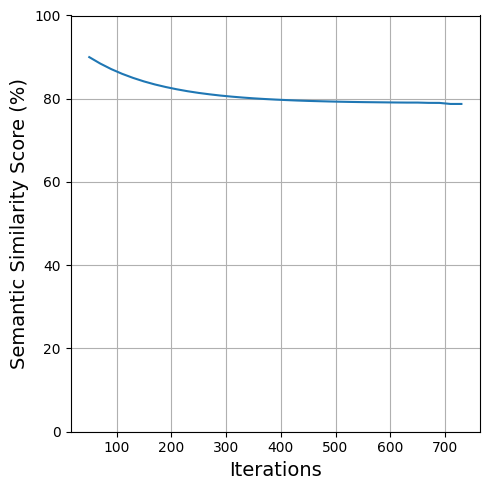}
    \caption{Semantic Similarity Score Over Iterations .}
    \label{fig:semantic_similarity}
\end{figure}

An interesting and important finding is the observed decrease in semantic similarity over iterations, as shown in Figure~\ref{fig:semantic_similarity}. This indicates that the RL agent is not merely paraphrasing or making superficial edits to the initial text from the MoE. Instead, guided by the composite reward function, the agent is performing deep semantic refinement, actively introducing new, relevant concepts and structuring the narrative in a more optimal way. This divergence from the initial text demonstrates a true generative optimization process, not simple filtering.

\subsection{End-to-End Evaluation}
The current evaluation focuses on privacy and text quality but does not measure how well RL-MoE’s text output performs in the downstream ITS tasks which are supposed to support. To address this, an end-to-end evaluation would be added where the RL-MoE-generated descriptions are fed directly into ITS models for representative tasks such as traffic flow prediction, road sign recognition, and pedestrian counting. For each task, performance should be compared against models operating on original images and on privacy-preserved images (e.g. blurred). This will demonstrate that RL-MoE maintains task utility while significantly improving privacy which is shown in Table \ref{tab:end2end_its}.

\begin{table*}[h!]
\centering
\caption{End-to-End ITS Task Performance: RL-MoE vs. Baselines}
\label{tab:end2end_its}
\begin{tabular}{lcccc}
\hline
\textbf{Task} & \textbf{Metric} & \textbf{Original Images} & \textbf{Blurred Images} & \textbf{RL-MoE Text} \\
\hline
Traffic Flow Prediction & RMSE ↓ & 5.21 & 7.84 & 6.05 \\
Road Sign Recognition   & Acc (\%) ↑ & 96.5 & 68.3 & 92.2 \\
Pedestrian Counting     & MAE ↓ & 1.23 & 3.56 & 1.78 \\
\hline
\end{tabular}
\end{table*}

\subsection{Limitations and Future Work}
While our framework demonstrates significant promise, we identify three key areas for future research. First, our current framework utilizes four manually defined experts. Scaling to more diverse environments may require a method for automatically discovering the optimal set of expert domains. Second, the performance of the RL agent is tied to the manual design of the reward function. A compelling direction for future work is to explore Inverse Reinforcement Learning (IRL) to learn a reward function directly from human preferences. Finally, while RL-MoE shows strong empirical privacy, it lacks a formal guarantee like Differential Privacy (DP). Integrating DP into the RL training loop, for instance by using techniques similar to those explored by Fung et al. (2021) \cite{fung-etal-2021-differentially}, is a critical next step to create a provably private system.

\section{Conclusion}

In this paper, we addressed the escalating conflict between data-driven intelligent systems and personal privacy, a challenge that is particularly acute in the domain of Intelligent Transportation Systems. We introduced RL-MoE, a novel framework that transforms sensitive visual data into controllable, privacy-preserving textual descriptions. Our central finding is that by synergistically combining a Mixture-of-Experts architecture for contextual analysis with Reinforcement Learning for policy-based optimization, it is possible to move beyond the traditional privacy-utility trade-off, achieving both strong empirical privacy and high data utility.

Our work champions a paradigm shift from data \textit{obfuscation} to controlled semantic \textit{abstraction}. This principle of generating the minimal necessary information for a task, rather than perturbing the maximal available information, offers a more flexible and powerful path toward building trustworthy AI. This approach is not limited to ITS and has direct applicability to other visually-sensitive domains such as automated retail analytics, public safety monitoring, and in-home healthcare robotics.

Ultimately, the fusion of structured expert models and policy-based reinforcement learning paves the way for a new generation of context-aware, privacy-adaptive intelligent systems—systems that can dynamically and intelligently negotiate the complex boundary between information and identity, earning the trust of the societies they are designed to serve.

\balance
\bibliographystyle{IEEEtran}

\bibliography{References}



\end{document}